\documentclass[letterpaper, 10 pt, conference]{ieeeconf}  

\IEEEoverridecommandlockouts                              
\usepackage[utf8]{inputenc}
\usepackage[T1]{fontenc}
\usepackage{url}
\usepackage{siunitx}
\usepackage{subcaption}
\usepackage{graphicx}
\usepackage{svg}
\usepackage{tikz}
\usepackage{breakcites}
\usepackage{amsmath,amssymb,amsfonts}
\usepackage{booktabs}
\usepackage{comment}
\makeatletter               
\let\NAT@parse\undefined    
\makeatother   
\usepackage[hyphenbreaks]{breakurl}
\newtheorem{assumption}{Assumption}

\title{\LARGE \bf
Predicting the Influence of Adverse Weather on Pedestrian Detection with Automotive Radar and Lidar Sensors
}

\author{Daniel Weihmayr$^{1, 2}$, Fatih Sezgin$^{1}$, Leon Tolksdorf$^{1, 3}$, Christian Birkner$^{1}$, Reza N. Jazar$^{2}$
\thanks{$^{1}$Technische Hochschule Ingolstadt, CARISSMA Institute of Safety in Future Mobility (C-ISAFE), Ingolstadt, Germany, e-mail: \{daniel.weihmayr, fatih.sezgin, leon.tolksdorf, christian.birkner\}@thi.de}%
\thanks{$^{2}$School of Mechanical and Automotive Engineering, Royal Melbourne Institute of Technology, VIC 3083 Australia, e-mail: s3933265@student.rmit.edu.au reza.nakahiejazar@rmit.edu.au}%
\thanks{$^{3}$Department of Dynamics and Control, Eindhoven University of Technology,
Eindhoven, The Netherlands, e-mail: l.t.tolksdorf@tue.nl}%
}

\newcommand\copyrighttext{%
  \footnotesize \textcopyright 2024 IEEE.  Personal use of this material is permitted.  Permission from IEEE must be obtained for all other uses, in any current or future media, including reprinting/republishing this material for advertising or promotional purposes, creating new collective works, for resale or redistribution to servers or lists, or reuse of any copyrighted component of this work in other works.}
\newcommand\copyrightnotice{%
\begin{tikzpicture}[remember picture,overlay]
\node[anchor=south,yshift=10pt] at (current page.south) {\fbox{\parbox{\dimexpr\textwidth-\fboxsep-\fboxrule\relax}{\copyrighttext}}};
\end{tikzpicture}%
}

\begin{document}

\maketitle
\thispagestyle{empty}
\pagestyle{empty}

\copyrightnotice
\begin{abstract}
Pedestrians are among the most endangered traffic participants in road traffic. While pedestrian detection in nominal conditions is well established, the sensor and, therefore, the pedestrian detection performance degrades under adverse weather conditions. Understanding the influences of rain and fog on a specific radar and lidar sensor requires extensive testing, and if the sensors' specifications are altered, a retesting effort is required. These challenges are addressed in this paper, firstly by conducting comprehensive measurements collecting empirical data of pedestrian detection performance under varying rain and fog intensities in a controlled environment, and secondly, by introducing a dedicated \textit{Weather Filter} (WF) model that predicts the effects of rain and fog on a user-specified radar and lidar on pedestrian detection performance. We use a state-of-the-art baseline model representing the physical relation of sensor specifications, which, however, lacks the representation of secondary weather effects, e.g., changes in pedestrian reflectivity or droplets on a sensor, and adjust it with empirical data to account for such. We find that our measurement results are in agreement with existent literature related to weather degredation and our WF outperforms the baseline model in predicting weather effects on pedestrian detection while only requiring a minimal testing effort.  
\end{abstract}

\section{Introduction}\label{sec:introduction} 
Accidents between manually controlled vehicles and vulnerable road users (VRUs), e.g., pedestrians and cyclists, are disproportionately high in rain and fog \cite{Andras.2021}. Adverse weather affects road users in two main ways: the road friction is decreased, and the visibility is reduced. Both effects apply in a similar way to automated vehicles (AVs). However, AVs are equipped with perception sensors such as cameras, lidars, and radars. These sensors capture essential information about the environment, including objects (e.g., pedestrians, cars, cyclists) and their characteristics. It provides the basis for decisions such as obstacle avoidance or emergency braking, emphasizing the importance of robustness under different environmental conditions. Many publications, therefore, show the problems that arise with the effects of rain and fog on the perception of the environment \cite{VargasRivero.2021, Vriesman.2020, Goelles.2020, Montalban.2021, Steinhauser.2021, Sezgin.2023}.\\
Due to the infinite number of possible road traffic scenarios, statistically certifying the safety of advanced driver assistance systems and AVs by only real-world testing becomes almost impossible \cite{SiddarthaKhastgir.2021}. Testing in simulation environments is becoming increasingly important, as many scenarios can be tested in a much shorter time \cite{Dona.2022}. However, this raises the challenge of designing a simulation that accurately reflects real-world phenomena, such as weather influences, which distort the perception of the sensors. While an empirical model for a specific sensor can be deduced from testing, extensive retesting may be required if the sensor specifications are changed.  
Here, comprehensive empirical testing data for radar and lidar sensors in a controlled rain and fog environment is sparse due to the need for large testing facilities and equipment allowing for reproducibility \cite{Zhang.2023}.
Further, a generalizable model that predicts weather effects on VRU detection for radar and lidar, independent of the sensor's specifications, is still an open research challenge.\\  
Rain and fog influence radar and lidar sensors in various ways. Here, they are categorized into two classes: primary and secondary effects. The primary effect is the attenuation caused by rain and fog, i.e., the received signal's power is lower than in nominal weather conditions \cite{Zang.2019}. Secondary effects are various influences combined, e.g., rain droplets on the sensor, clutter, backscatter, and changes in target reflectivity, amongst others \cite{Winner.2016}. State-of-the-art models are often only capturing primary weather effects (e.g., \cite{Li.2021}, \cite{Du.2004}, \cite{Ashraf.2018}). Models considering secondary effects and including validation with real measurements are sparse \cite{Haider.2023, Kettelgerdes.2023, Hahner.2021}. Modeling primary weather effects requires a deep understanding of the weather composition and the subsequent impact on the specific sensor. Clearly, that approach lacks secondary weather effects, as the complexity and impact on the specific sensor are challenging to determine with analytical models \cite{Zang.2019}. The advantage of primary modeling is that weather effects can be calculated independently of the scenario and its correlation with the sensor hardware used. In contrast, a purely empirical approach involves performing extensive measurements with a specific set of sensors in a predefined set of scenarios. While this approach captures all measured primary and secondary effects, it lacks the ability to generalize beyond the measured scenarios and the sensor hardware used, as it is unclear if the measured effects are caused by the weather or the specific sensor.\\
We present a method combining both approaches to derive the so-called \textit{Weather Filter} (WF) model. The WF models the degradation of the pedestrian detection performance of radar and lidar as if operating in user-specified rain and fog levels. Therefore, we use a baseline model that represents the primary weather effects and adapt the model with empirical data obtained from comprehensive measurements, including the secondary effects. As such, we obtain the WF, improving the prediction accuracy of the maximum detection range for a pedestrian in user-specified rain and fog compared to the baseline model. Further, we show that only a limited number of measurements are needed to adjust the WF to specific sensor hardware. Precisely, we contribute: 1. comprehensive measurements of the pedestrian detection performance for radar and lidar sensors under reproducible rain and fog conditions, that we find to be in agreement with existing literature, 2. adjusting a state-of-the-art model with empirical data to account for secondary weather effects while allowing for user-defined sensor specifications, which we find to be outperforming the state-of-the-art model. \\
The remainder of this article is organized as follows: The the problem is defined in Section \ref{sec:problem_formulation}. Here, assumptions are stated, and weather modelling is described. Section \ref{sec:measurements} describes the measurements conducted, providing a reproducible and systematic variation. Section \ref{sec:weather_filter} introduces the WF to combine a baseline model with empirical data to improve prediction accuracy. Section \ref{sec:results} presents the measurement results and the predictions estimated by the baseline model compared to the WF. Section \ref{sec:discussion} discusses the outcomes and their interpretations. Lastly, Section \ref{sec:conclusion} briefly summarises the results and outlines future work for further research.

\section{Problem Formulation}\label{sec:problem_formulation}
As state-of-the-art models only represent the primary effects of rain and fog on radar and lidar, pedestrian detection performance, our objective is to empirically tune such a model to account for the shortfalls, i.e., the lack of representation of secondary weather effects.
\begin{assumption}\label{ass:detection}
 Suppose that $m_R$ target-related and recurring detection points are necessary for an object to be detected by a radar and $m_L$ for a lidar. Further, the number of recurring target-related detection points $m$ over multiple measurement frames $i \in \mathbb{N}$ is proportional to the detection distance $d \in \mathbb{R}$, i.e., $i \propto d$.
\end{assumption}
\begin{figure}[h]
\centering
\begin{subfigure}[b]{0.387\linewidth}
          \includegraphics[width=\linewidth]{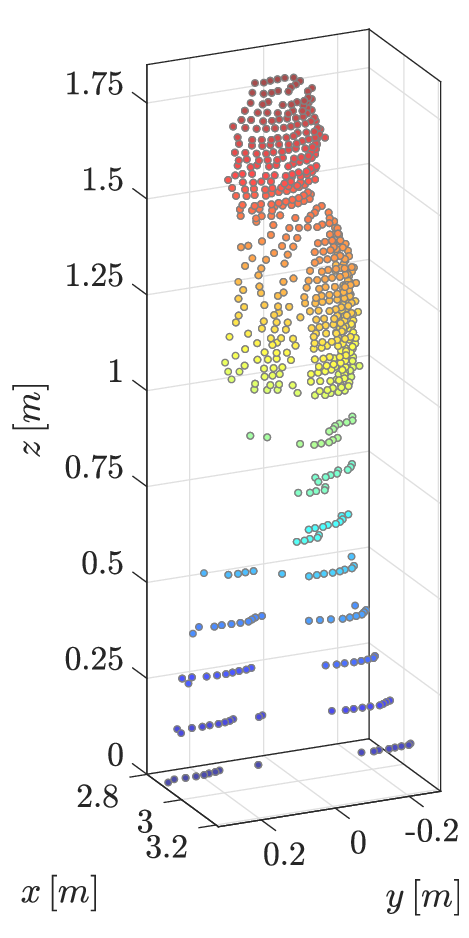}
          \captionsetup{font=footnotesize}
            \caption{Angled view}
   \end{subfigure}
    \begin{subfigure}[b]{0.321\linewidth}
           \includegraphics[width=\linewidth]{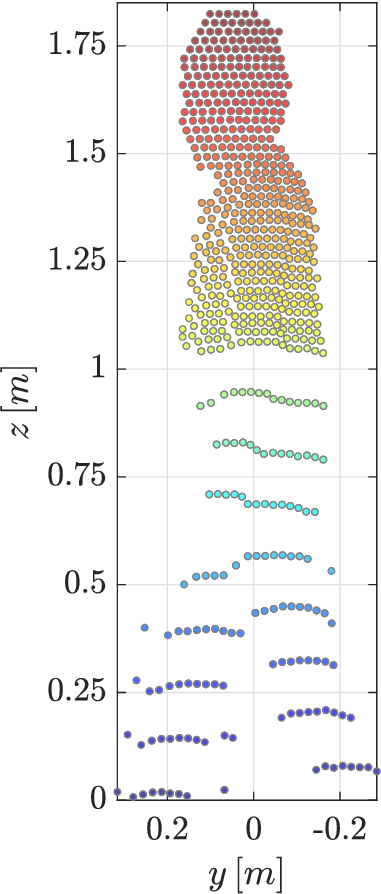}
           \captionsetup{font=footnotesize}
             \caption{Side view}
    \end{subfigure}
    \begin{subfigure}[b]{0.263\linewidth}
           \includegraphics[width=\linewidth]{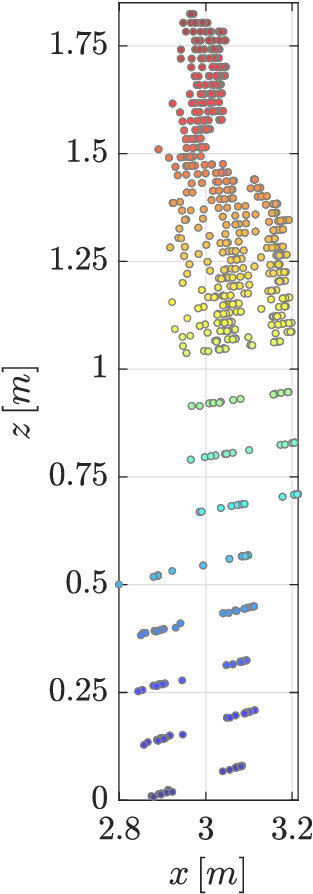}
           \captionsetup{font=footnotesize}
             \caption{Front view}
    \end{subfigure}
    \captionsetup{font=footnotesize}
\caption{Recurring lidar detection points on a pedestrian dummy.}
\label{fig:lidar_pedestrian_point_cloud}
\end{figure}
\begin{figure}[h!]
 \centering
 \begin{subfigure}[b]{0.49\linewidth}
           \includegraphics[width=\linewidth]{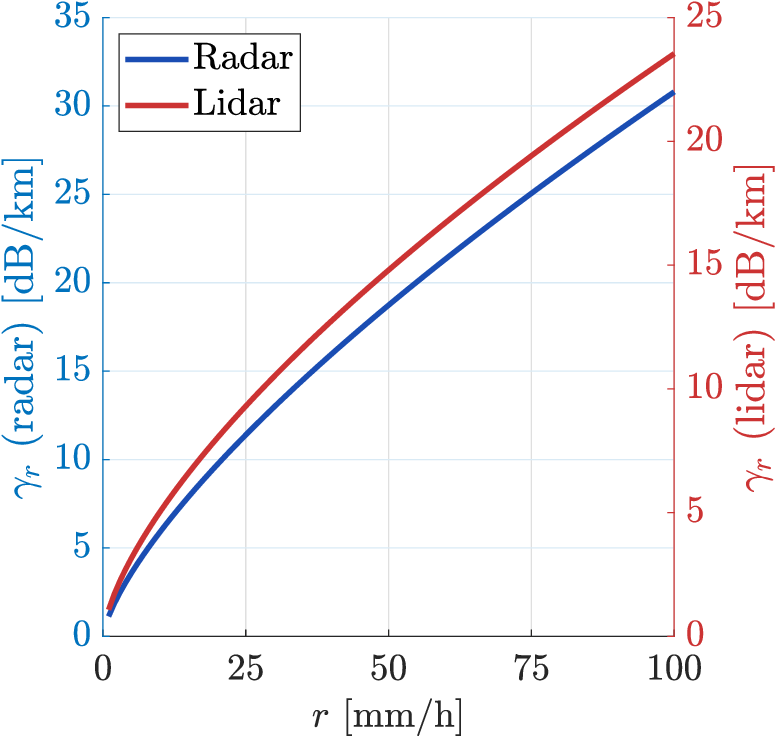}
    \end{subfigure}
     \begin{subfigure}[b]{0.49\linewidth}
            \includegraphics[width=\linewidth]{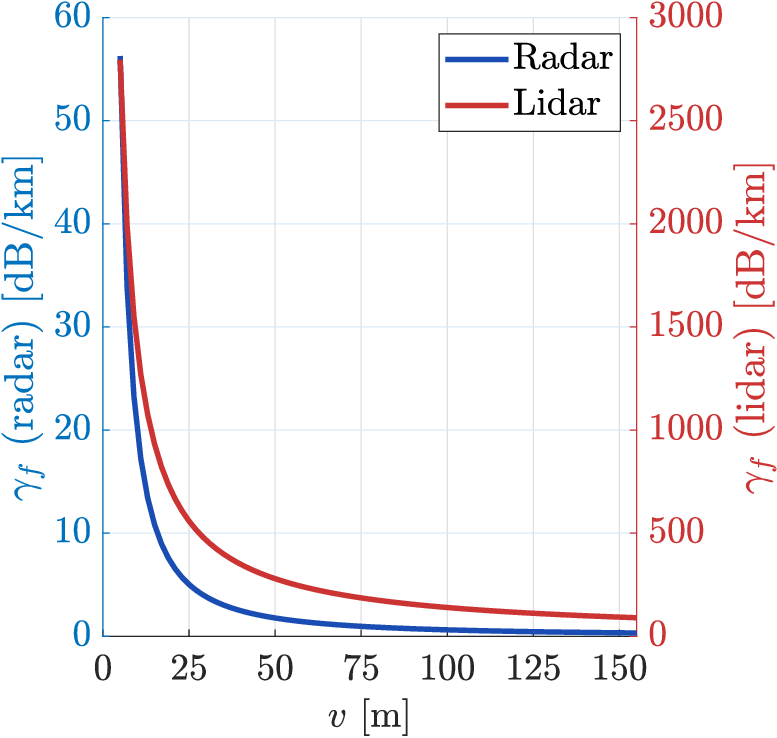}
     \end{subfigure}
     \captionsetup{font=footnotesize}
 \caption{Baseline model of $\gamma_r$ and $\gamma_f$ over rain and fog rate for radar and lidar.}
 \label{fig:combined_attenuation_rain_fog}
 \end{figure}
This assumption implies that variations in attenuation due to adverse weather conditions directly influence the sensor's ability to generate a sufficient number of consistent detection points, thereby impacting the overall object detection performance. An example for recurring detection points on a pedestrian dummy is provided in Figure \ref{fig:lidar_pedestrian_point_cloud}. Following Assumption \ref{ass:detection}, we denote the average number of recurring detection points over all measurement frames per distance, with $\bar{N}_{d}$. Frames contain recurring measurement points in $\mathbb{R}^3$ for a discrete time instance, compromising a preceding data processing at the operating frequency to obtain only recurring measurement points. Given $N$ frames and $N_{i,d}$ recurring detection points per frame and distance, then for 
\begin{equation} \label{eq:detection_condition}
\bar{N}_{d} = \frac{1}{N} \sum_{i=1}^{N} N_{i,d},
\end{equation}
a target is said to be detected at a specific position if the following conditions are satisfied for radar and lidar, respectively,
\begin{equation} \label{eq:detection_condition}
\bar{N}_{R,d} \geq m_R, \qquad \bar{N}_{L,d} \geq m_L,
\end{equation}
with the subscript $R$ denoting radar and $L$ denoting lidar. Further, we assume that different forms of attenuation can comprehensively represent each sensor's primary degradation on the pedestrian detection performance. Let the overall attenuation $\gamma$ be expressed as a combination of individual attenuations for rain $\gamma_r$, fog $\gamma_f$, and the atmosphere $\gamma_a$ as:
\begin{equation} \label{eq:gamma_combined_new}
\gamma = \gamma_r + \gamma_f + \gamma_a.
\end{equation}
While the attenuation for rain, fog, and the atmosphere for primary weather effects can be estimated with various methods \cite{Du.2004}, \cite{Li.2021}, \cite{Ashraf.2018}, it is challenging to do so for the disturbances caused by secondary weather effects. To compromise each individual attenuation into a model, let the attenuation of rain $\gamma_{R,r}$ and fog $\gamma_{R,f}$ on a radar sensor further be given by: 
\begin{equation} \label{eq:gamma_r_f}
\gamma_{R,r} = \eta_{R,r} \, k_R \, r^{\alpha_R}, \qquad \gamma_{R,f} = \eta_{R,f} \, b_R \, M,
\end{equation}
with modeling parameters $k_r, b_R, \alpha_R$, the rain rate $r$ for the radar, and the extension with the introduced empirical tuning coefficients $\eta_{R,r}$, $\eta_{R,f}$. Note that with $\eta_{R,r} = 1$ and $\eta_{R,f} = 1$, the baseline model from \cite{Itu.2005, Morabito.2014} is retrieved. Further, the visual range $v$ is described by the fog density $M$ as well as the fog type parameter $c_f$ (see \cite{Muhammad.2010}) for dry continental fog, 
\begin{equation} \label{eq:fog_density}
M = \left({\frac{c_f}{v}}\right)^{{\frac{3}{2}}}.
\end{equation}

The effects of rain on a lidar sensor are similar to (\ref{eq:gamma_r_f}), where tuning parameters $k_L$ and $\alpha_L$ for rain attenuation are given in \cite{Carbonneau.1998}. Thus, the attenuation of fog on a lidar sensor is modeled as:

\begin{equation} \label{eq:gamma_f_lidar}
\gamma_{L,r} = \eta_{L,r} \,k_L \, r^{\alpha_L}, \qquad  \gamma_{L,f} = \eta_{L,f} \,{\frac{17}{v}} \,\biggl({{\frac{\lambda_L}{\lambda_0}\biggl)^{-q}}},
\end{equation}

where $\lambda_L$, $\lambda_{0}$, $q$ represents the lidar's wavelength, the reference wavelength, and the absorption coefficient, respectively. Additionally and similar to (\ref{eq:gamma_r_f}), we introduce the the empirical tuning coefficients $\eta_{L,r}$ and $\eta_{L,f}$. With $\eta_{L,r} = 1$ and $\eta_{L,f} = 1$ the baseline model from \cite{Fiorino.2008} is retrieved. The behavior of the \textit{baseline model}, i.e. (\ref{eq:gamma_r_f}) - (\ref{eq:gamma_f_lidar}) with the empirical tuning coefficients set to one, is depicted in Figure \ref{fig:combined_attenuation_rain_fog}, based on the values according to Table \ref{tab:attenuatino_parameters}.\\
\textit{Problem:} Given Assumption \ref{ass:detection} and Condition (\ref{eq:detection_condition}), the goal is to empirically adjust the introduced tuning coefficients $\eta_{R,r}$, $\eta_{R,f}$, $\eta_{L,r}$, and $\eta_{L,f}$ to account for the secondary weather effects on pedestrian detection performance.

\section{Measurements}\label{sec:measurements}
The exact influence of weather conditions on vehicle sensors is difficult to determine, as reproducibility is a particular challenge during measurements. In order to improve the consistency of the measurements, artificial rain, and fog are used in the indoor test facility of CARISSMA \cite{Hasirlioglu.2017, Hasirlioglu.2016}. The artificially generated weather conditions' reproducibility plays an overriding role compared to the difference between artificial and natural rain and fog \cite{Hasirlioglu.2020}.

\begin{figure}[h!]
	\centering
	\includegraphics[width=0.99\linewidth]{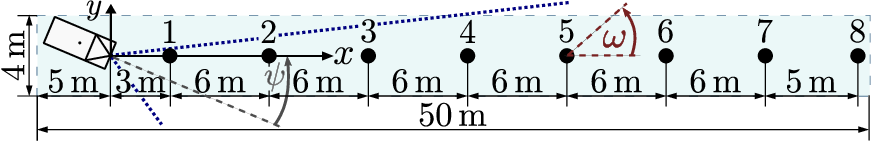}
 \captionsetup{font=footnotesize}
	\caption{Measurement setup for CARISSMA indoor weather testing.}
	\label{fig:test_setup}
\end{figure} 
The indoor test facility offers a test area of $1260\,\si{\meter}^2$ for fog tests. In addition, an area of $50 \times 4$ meters overlapping with this area is available for rain tests. Figure \ref{fig:test_setup} shows that discrete measurements are carried out at a defined distance from the test vehicle with a stationary adult pedestrian dummy \cite{4activeSystems.2021} to ensure consistency. The test vehicle with a radar and a lidar can be rotated by the angle $\psi$ at the origin to investigate the angle-dependent detection range. Further, $\omega \in [0,2\,\pi[$ denotes the pedestrian's rotation angle. The pedestrian is positioned on eight points at a distance $d_p \in \{3, 9, 15, 21, 27, 33, 39, 44\} \, \si{\meter}$ from the origin, oriented such as it is crossing the road from right to left, i.e., $\omega = 0.5\,\pi$. Note that a measurement without the pedestrian dummy is also recorded for each weather condition. The measured rain conditions are $r\in \{0, 16, 98\}$\,\si{\milli\meter/\hour}, where \qty{0}{\milli\meter/\hour} represents a measurement without rain and \qty{98}{\milli\meter/\hour} the strongest rain offered by the rain system. Respectively, the visual ranges for the fog measurements are $v\in \{6,20, \infty\}$\,\si{\meter}. Here, $\infty$\,\si{\meter} represents a setting without fog and $6$\,\si{\meter} the densest fog the test facility can generate. The selected weather intensities are set to expose the sensors to severe weather conditions. Further, the rain intensity of \qty{16}{\milli\meter/\hour}, which is already considered as heavy rain \cite{Jebson.2007}, and fog with wider visibility, i.e., \qty{20}{m}, which is defined as dense fog \cite{Ali.2024}, are included as an intermediate sampling point. For the selected fog visibility ranges, the focus is particularly on the challenge of detecting pedestrians in the near field. While the visibility in fog is exactly specified, we note that it fluctuates due to the cloudy nature of fog, with the resulting measurement boundaries of $\SIrange{5}{7}{\meter}$ and $\SIrange{17}{23}{\meter}$. The visibility in fog is measured with the SICK VISIC620 measuring device.  
Figure \ref{fig:lidar_pedestrian_point_cloud} shows the lidar point cloud of the pedestrian dummy. Note that the top mounted lidar's vertical resolution is less dense in the near field ($d<5$\,\si{\meter}) to increase the limited vertical resolution in the mid-to far-range, explaining the fewer number of detection points in the lower part of the pedestrian. The duration of measurements is set to five seconds for both sensors and each measurement. In the case of radar, only recurring detection points are included in the detection list to avoid a distorted number of target-related detections due to reflections from the rainfall (backscatter).

\section{Weather Filter}\label{sec:weather_filter}
An open research challenge we address is the applicability of our weather model for differently specified lidar and radar sensors. The WF predicts the maximum detection range for a pedestrian for a specified radar and lidar sensor under user-defined rain and fog conditions. Such reduces the need to test individual sensors to adjust sensor simulation environments. Therefore, however, we require an integration of individual sensor specifications. 
Table \ref{tab:nomenclature} provides the definition of considered symbols.
\begin{table}[h]  
\captionsetup{font=footnotesize}
   \caption{Nomenclature of sensor and environment specifications.}
   \small
   \centering
   \begin{tabular}{cr|cr}
   \toprule\toprule
   \textbf{Name} & \textbf{Meaning} & \textbf{Name} & \textbf{Meaning} \\ 
   \midrule
 $A$ & Surface & $G$ & Antenna gain \\
$h$ & Target height & $l$ & Target length \\
$P_n$ & Detection threshold & $P_o$ & Received power \\
$P_t$ & Transmitted power & $\sigma$ & Standard deviation \\
$\theta$ & Environment temp. & $T$ & Transmission \\
$Q_h$ & Horizontal divergence & $Q_v$ & Vertical divergence \\
$w$ & Target width & $\Gamma$ & Detection range \\
$\omega$ & Target rotation & $\varPhi$ & Reflection angle \\
$\varrho$ & Object reflectance & $\varsigma$ & Radar cross section \\
$\xi_R$ & Offset calibration \\
   \bottomrule  
   \end{tabular}
   \label{tab:nomenclature}
\end{table}

Our approach is to create a virtual model of the sensors to be tested based on their individual sensor specifications in combination with the results of limited discrete measurements in order to parameterize calibration factors in  (\ref{eq:gamma_r_f}) and (\ref{eq:gamma_f_lidar}).

\subsubsection{Radar}

For the radar, the correlation between the transmitted and received power, with regard to the maximum range per target parameters, is given by \cite{Skolnik.2008}: 
\begin{equation} \label{eq:radar_range}
P_{o} = 10^{\frac{-\gamma \, \Gamma	}{1000}} \, {\frac{P_t \, \xi_R\, G^2 \, \varsigma \, \lambda_R^2}{4\pi^3 \, \Gamma^4}}.
\end{equation}
Here, ideal conditions are presumed, wherein the propagation of electromagnetic waves via an isotropic spherical radiator is assumed.

\subsubsection{Lidar}
Similar to the radar, the correlation between transmitted and received power of the lidar is provided by \cite{Winner.2015}: 
\begin{equation} \label{eq:lidar_range}
P_{o} = 10^{\frac{-\gamma \, \Gamma	}{1000}} \, {\frac{\varrho \, A_l \, w \, T^2 \, P_t}{\pi^2 \, \Gamma^4 \, {\frac{Q_v \, Q_h}{4}} \, \Bigl({\frac{\varPhi}{2}}\Bigl)^2 }}.
\end{equation}
In order to retrieve the detection range for an individual detection point, we use a grid search to solve (\ref{eq:radar_range}) and (\ref{eq:lidar_range}) for $\Gamma$. An individual point is considered to be detected if the condition $P_o \geq P_n$, to account for each sensor's minimum detection threshold $P_n$ is satisfied. For our model, we specify the sensors according to Table \ref{tab:radar_lidar_simulation_parameters}.
Additionally, the calibration factors  $\eta_{R}$ and $\eta_{L}$ are tuned using the number of target-related detection points at $d$ and the furthest empirically measured distance of $d_p$ that fulfills the condition imposed by (\ref{eq:detection_condition}). If, e.g., $d_{2}<m_R$ or $d_{2}<m_L$, only the number of detection points at $d_{1}$ is selected for the calibration, where $\square_{p}$ represents the position according to Figure \ref{fig:test_setup} with $p\in [1, ..., 8]$. The parameterization is carried out using nonlinear regression to minimize the distances between $\Gamma$ and $d$ with the optimization variable $\eta$ under Assumption \ref{ass:detection}. 
Note that increasing the number of calibration factors results in an increasing reliance on the empirical data and a gradual loss of the fundamental physical structure, leading to overfitting.

\section{Results} \label{sec:results}
The sensor specification, used for the empirical quantification of weather effects on the pedestrian detection, is given in Table \ref{tab:radar_lidar_simulation_parameters} for radar and lidar, respectively.

\begin{table}[h]    
\captionsetup{font=footnotesize}
    \caption{Sensor specifications with values for radar in the upper part and lidar in the lower part.} 
    \small
    \centering
    \begin{tabular}{cr|cr}
    \toprule\toprule
    \textbf{Parameter} & \textbf{Value} & \textbf{Parameter} & \textbf{Value} \\ 
    \midrule
  $G$ & $16 \, \si{\decibel i}$ & $m_R$ & $1$ \\
  $P_n$ & $5 \times 10^{-12} \, \si{\watt}$ & $P_t$ & $1 \times 10^{-2} \, \si{\watt}$\\
  $\gamma_a$ & $0.6 \, \si{\decibel}$& $1/\lambda$ & $77 \, \si{\giga\hertz}$ \\
  $\xi_R$ & $1,875$ \\
\midrule
 $A_l$ & $4.4 \times 10^{-2} \, \si{\meter}^2$ & $H$ & $0.5 \, \si{\meter}$ \\
$m_L$ & $10$ &$P_n$ & $1 \times 10^{-8} \, \si{\watt}$\\
  $P_t$ & $22 \times 10^{-2} \, \si{\watt}$ & $Q_h$ & $18.27 \, \si{\milli\radian}$ \\
  $Q_v$ & $4.57 \, \si{\milli\radian}$ &$T$ & $0.9$\\
  $\eta_{L,f}$ & $0.199$ &$\eta_{L,r}$ & $1.063$\\
  $\gamma_a$ & $0.03 \, \si{\decibel}$ & $\lambda_L$ & $905 \times 10^{-9} \, \si{\meter}$\\
 $\varPhi$ & $0.5 \, \pi$ \\
    \bottomrule  
    \end{tabular}
    \label{tab:radar_lidar_simulation_parameters}
 \end{table}

\subsection{Empirical Weather Impact}\label{sec:results_measurements}

\begin{figure}[h!]
	\centering
	\includegraphics[width=0.99\linewidth]{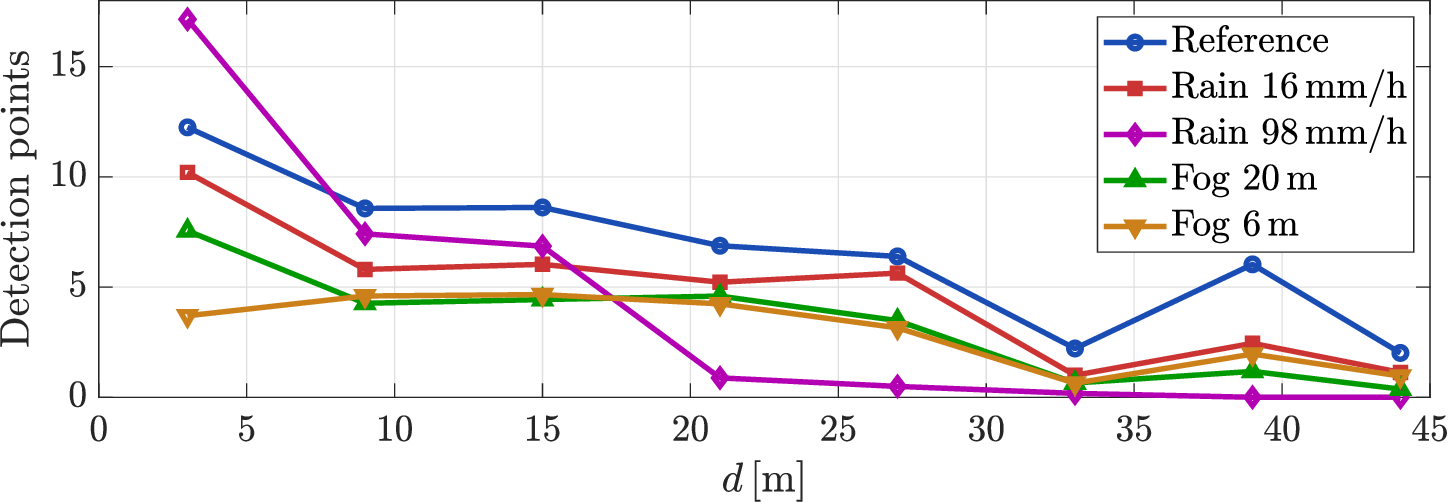}
 \captionsetup{font=footnotesize}
	\caption{Number of target related and recurring radar detection points for a pedestrian under rain and fog. Reference denotes a measurement without rain or fog.}
	\label{fig:detection_points_radar}
\end{figure}

\begin{figure}[h!]
	\centering
	\includegraphics[width=0.99\linewidth]{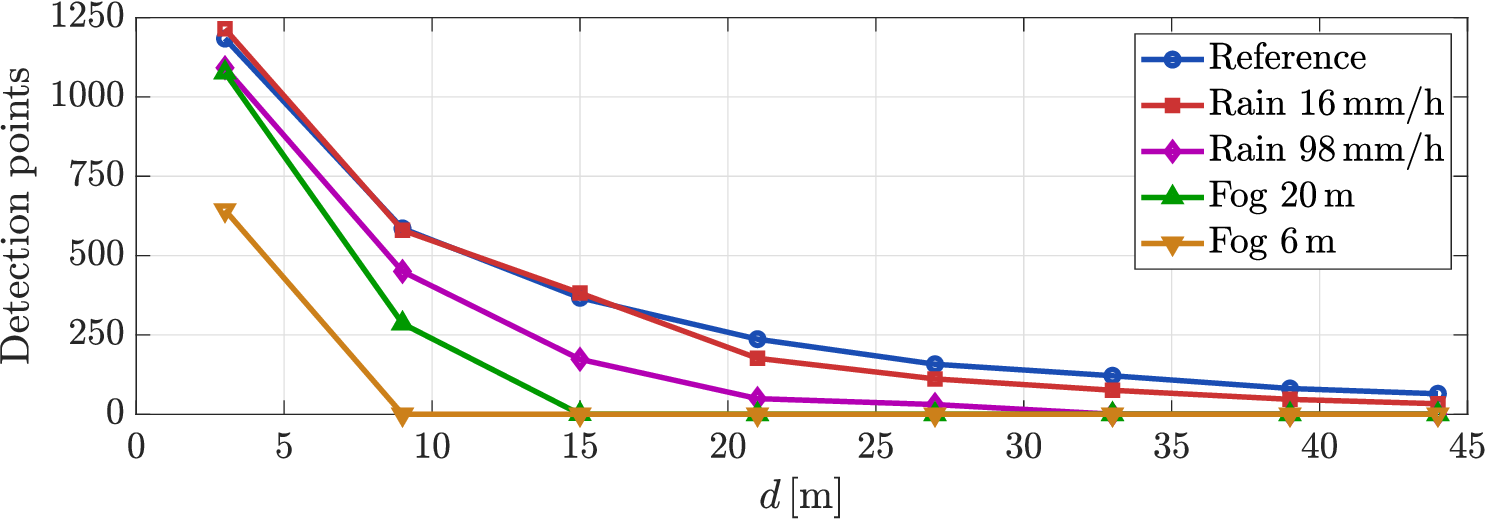}
 \captionsetup{font=footnotesize}
	\caption{Number of target related lidar detection points for a pedestrian under rain and fog. Reference denotes a measurement without rain or fog.}
	\label{fig:detection_points_lidar}
\end{figure}

The measurements are evaluated based on individual sensor detection points in the three-dimensional space, which are obtained by the conditions outlined in Section \ref{sec:weather_filter}.
Figures \ref{fig:detection_points_radar} and \ref{fig:detection_points_lidar} contain the number of recurring detection points on the pedestrian dummy for the respective sensors and weather conditions, providing the mean number of recurring detection points over the measurement period.
The number of pedestrian-related detection points includes a backscatter compensation for the lidar. In this process, a free-space measurement is conducted under identical weather conditions without the pedestrian. Subsequently, false-positive detections within the analyzed volume, excluding the pedestrian volume, are obtained. The volume is visualized in Figure \ref{fig:lidar_pedestrian_point_cloud} with a pedestrian dummy included. Backscatter compensation is not used for the radar, as this particular radar shows largely robust behavior in free-space weather measurements. 
Figure \ref{fig:detection_points_radar}, however, shows a higher number of detection points than expected for $r = 98\,\si{\milli\meter/\hour}$ up to the distance $d = 15\,\si{\meter}$. Although the deployed radar only stores recurring detections, the heavy rain affects the number of detection points in the near field.

The following Figures, \ref{fig:sigma_radar} and \ref{fig:sigma_lidar}, show the standard deviation of the detections across all measurement frames and thus illustrate the dynamics of the measurement data. It should be noted that isotropic antennas are assumed, so the standard deviation is influenced by the spatial resolution of the transmitters, which depends on $d$. In addition, the component of horizontal and vertical beam divergence, as shown in Table \ref{tab:radar_lidar_simulation_parameters}, becomes relevant for lidar at greater distances.
\begin{figure}[h!]
	\centering
	\includegraphics[width=0.99\linewidth]{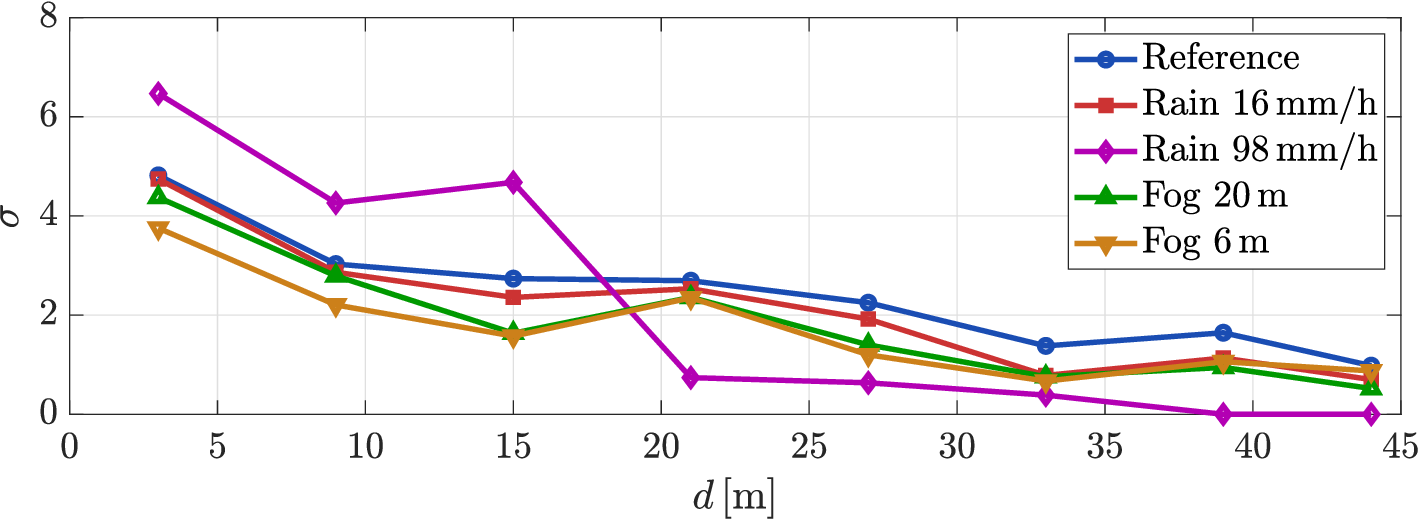}
 \captionsetup{font=footnotesize}
	\caption{Standard deviation $\sigma$ of the radar detection points over all individual frames per position.}
	\label{fig:sigma_radar}
\end{figure}
\begin{figure}[h!]
	\centering
	\includegraphics[width=0.99\linewidth]{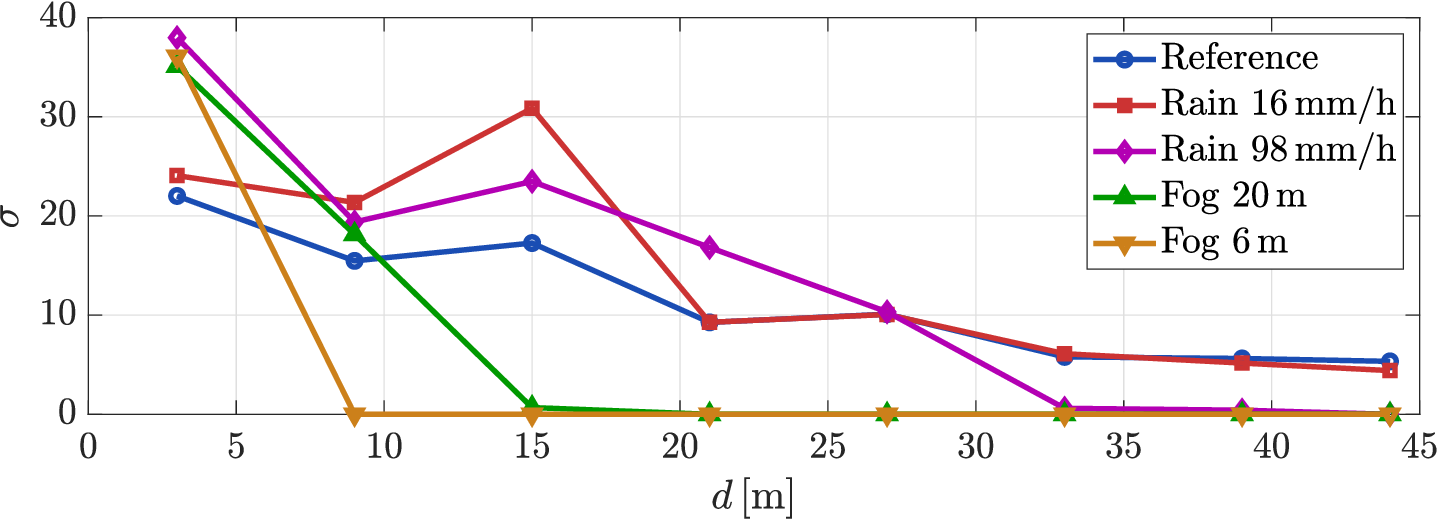}
 \captionsetup{font=footnotesize}
	\caption{Standard deviation $\sigma$ of the lidar detection points over all individual frames per position.}
	\label{fig:sigma_lidar}
\end{figure}
Analysis of the standard deviation for rain in Figures \ref{fig:sigma_radar} and \ref{fig:sigma_lidar} reveals a measurement outlier at $d = 15 \, $m. This anomaly is attributed to an irregular rainfall pattern specifically at that location. This outlier is not considered during the calibration of the WF.
Overall, the observed behaviour of the sensors in the measurements is generally consistent with the descriptions in existing literature \cite{Heinzler.2019, Kim.2021, Muckenhuber.2022}.

\subsection{Modeled Weather Impact}
From (\ref{eq:gamma_r_f}) - (\ref{eq:gamma_f_lidar}), we obtain the corresponding attenuations for radar and lidar in rain and fog, presented in Figure \ref{fig:combined_attenuation_rain_fog} with the values from Table \ref{tab:attenuatino_parameters}.
In relative terms, the radar experiences a slightly higher rain-induced degradation than the lidar. In contrast, the influence of fog on the lidar is significantly higher than radar.  
Further, we set the virtual environment's parameters\footnote{Note that weather effects on different types of target classes necessitate consideration of target related information as listed in Table \ref{tab:simulation_environment_parameters}.} according to Table \ref{tab:simulation_environment_parameters}, with $\varsigma$ retrieved from \cite{IPG.2020} and $c_f$ from \cite{Muhammad.2010}.

\begin{table}[h]    
\captionsetup{font=footnotesize}
    \caption{Attenuation parameters for lidar (left) and radar (right).} 
    \small
    \centering
    \begin{tabular}{cr|cr}
    \toprule\toprule
    \textbf{Parameter} & \textbf{Value} & \textbf{Parameter} & \textbf{Value} \\ 
    \midrule
  $k_L$ & $1,076$ & $b_R$ & $$3,1733$$ \\ 
  $q$ & $3,45 \times 10^{-2}$ & $k_R$ & $1,1319$ \\ 
  $\alpha_L$ & $0,67$ & $\alpha_R$ & $0,7174$ \\ 
  $\lambda_0$ & $550 \times 10^{-9} \, \si{\meter}$ & $\eta_{R,r}$ & $1,163$\\ 
  $\lambda_L$ & $905 \times 10^{-9} \, \si{\meter}$ & $\eta_{R,f}$ & $0,0199$\\ 
  $\eta_{L,r}$ &  $1,163$ &  &  \\ 
  $\eta_{L,f}$ &  $0,199$ &  &  \\ 

   \bottomrule  
    \end{tabular}
    \label{tab:attenuatino_parameters}
 \end{table}

\begin{table}[h]    
\captionsetup{font=footnotesize}
    \caption{Simulation environment parameters related to the pedestrian.} 
    \small
    \centering
    \begin{tabular}{cr|cr}
    \toprule\toprule
    \textbf{Parameter} & \textbf{Value} & \textbf{Parameter} & \textbf{Value} \\ 
    \midrule
  $A$ & $0,72 \, \si{\meter}^2$ & $\varrho$ & $0,5$ \\
   $w$ & $0,4 \, \si{\meter}$ & $l$ &  $0,3 \, \si{\meter}$\\
   $\omega$ &  $90 \si{\degree}$  & $\varsigma$ &  $10,08\, \si{\meter}^2$\\
   $h$ &  $1,8 \, \si{\meter}$ & $c_f$ &  $0,034$  \\
   $\vartheta$ &  $10 \, ^\circ\text{C}$   \\
    \bottomrule  
    \end{tabular}
    \label{tab:simulation_environment_parameters}
 \end{table}

\begin{figure}[h!]
 	\centering
 	\includegraphics[width=0.99\linewidth]{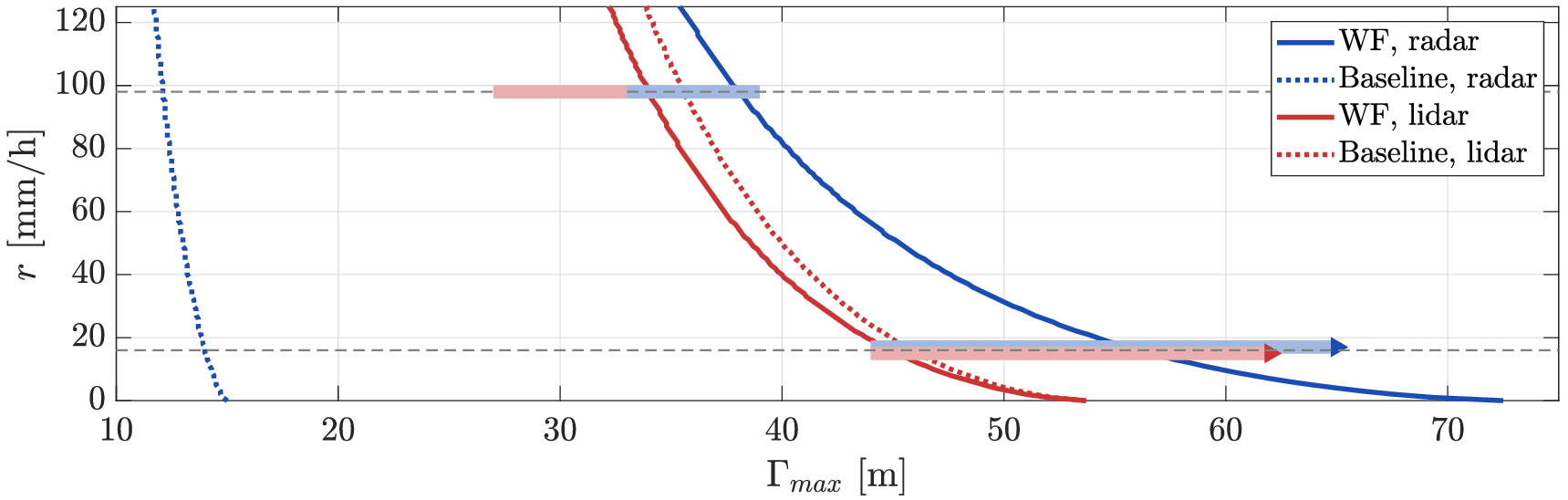}
  \captionsetup{font=footnotesize}
 	\caption{Detection threshold comparison between WF and baseline approach for radar and lidar under rain.}
 	\label{fig:combined_lidar_radar_rain}
 \end{figure}

\begin{figure}[h!]
 	\centering
 	\includegraphics[width=0.99\linewidth]{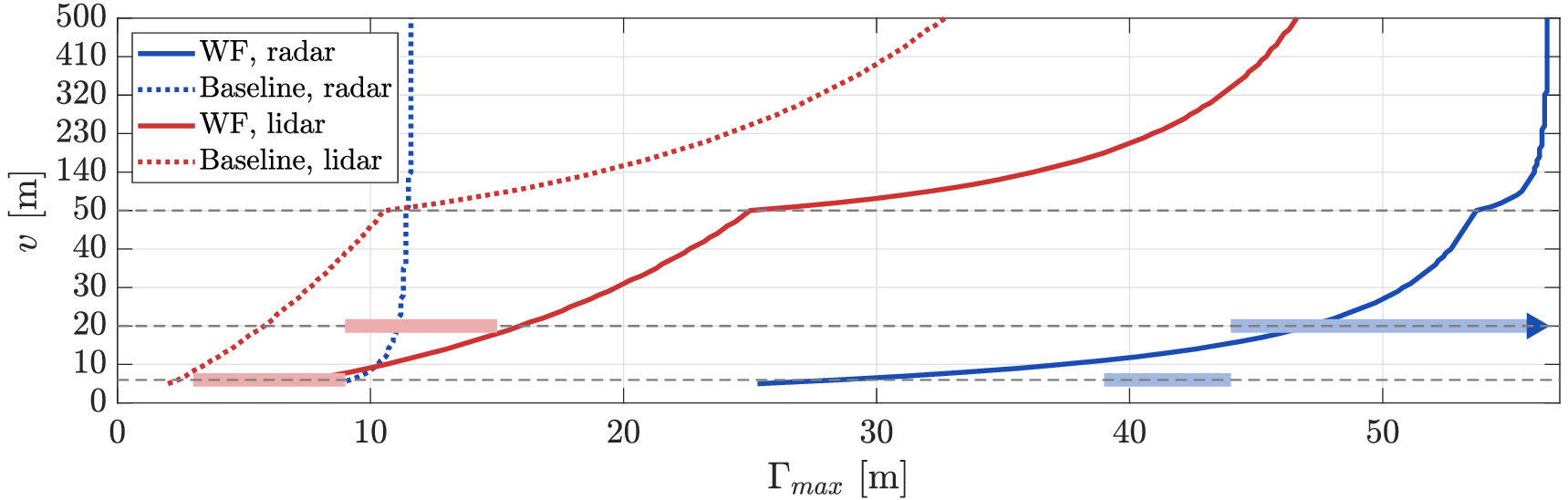}
  \captionsetup{font=footnotesize}
 	\caption{Detection threshold comparison between WF and baseline approach for radar and lidar under fog. The y-axis scales different from $v = 50 \,\si{\meter}$.}
 	\label{fig:combined_thresholds__radar_lidar_fog}
 \end{figure}

The correlation between the rain rate $r$ and the maximum detection range $\Gamma_{max}$ is depicted for both sensors in Figure \ref{fig:combined_lidar_radar_rain}, presenting the calibrated WF, as well as the baseline approach, i.e., $\eta_R = 1$, $\eta_L = 1$, and $\xi_R = 1$. Additionally, we plotted the measured maximum detection range between the discrete measurement positions from Figure \ref{fig:combined_lidar_radar_rain} with horizontal transparent lines indicating the limits and the arrow indicating an undefined end which is outside the scope of the testing capabilities of CARISSMA. For the lidar, it can be seen that the baseline approach already closely matches the WF's predictions, which is compensated by $\eta_{L,r}$. For the radar, a significant difference between the baseline approach and the empirical results is found equally for rain and fog (Figure \ref{fig:combined_thresholds__radar_lidar_fog}). This offset is independent of $\gamma$ and can be traced back to (\ref{eq:radar_range}). We introduce the parameter $\xi_R$ to compensate for this offset and adjust it accordingly based on the regression technique used for $\eta_R$ and $\eta_L$. 
Similarly, the correlation between $v$ and $\Gamma_{max}$ is presented in Figure \ref{fig:combined_thresholds__radar_lidar_fog}. The radar shows the same overall robustness against fog's secondary weather effects once enhanced by $\xi_R$. Subsequently, no further attenuation coefficient is needed, resulting in $\eta_{R,r} = 1$ and $\eta_{R,f} = 1$.

\begin{figure}[h!]
 	\centering
 	\includegraphics[width=0.99\linewidth]{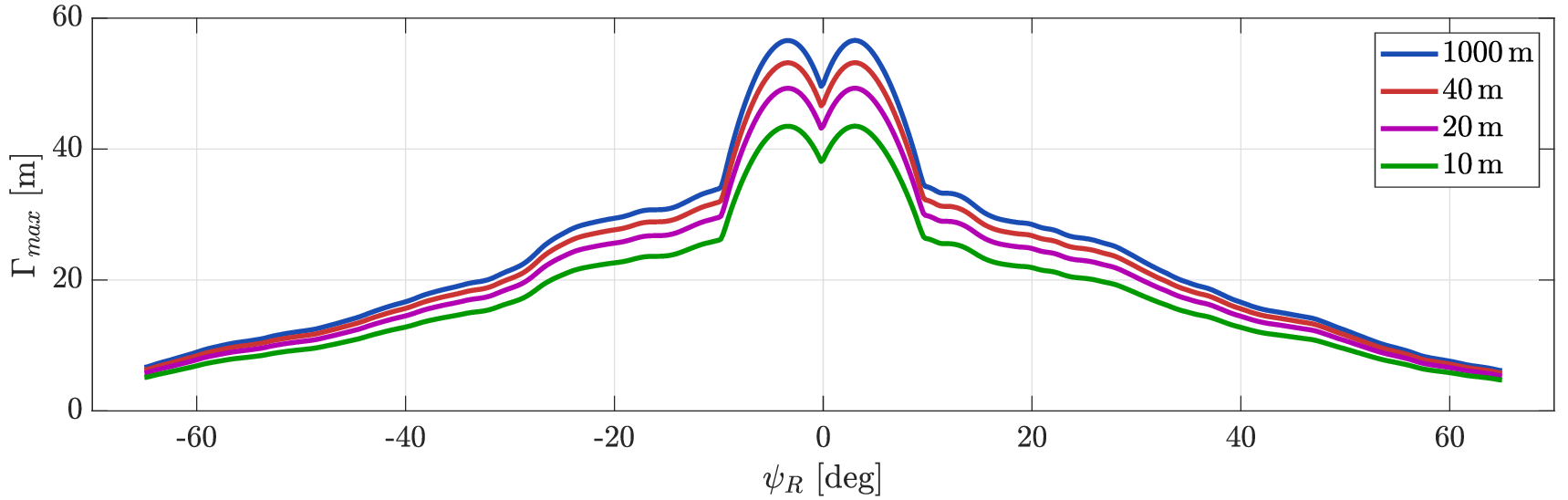}
  \captionsetup{font=footnotesize}
 	\caption{Radar detection range under different fog intensities for a pedestrian target.}
 	\label{fig:thresholds_radar_fog_azimuth}
 \end{figure}

The radar has no uniform field-of-view (FOV) but is subject to an angle-dependent transmit and receive power. The angular dependence of $\psi_R \in [-65,65]$° (see Figure \ref{fig:test_setup}) is a factor and can be regarded as consistent and weather-independent based on the manufacturer's specifications, which is confirmed by additional empirical measurements under different weather conditions \cite{Loeffler.2021}.
 \begin{figure}[h!]
 	\centering
 	\includegraphics[width=0.99\linewidth]{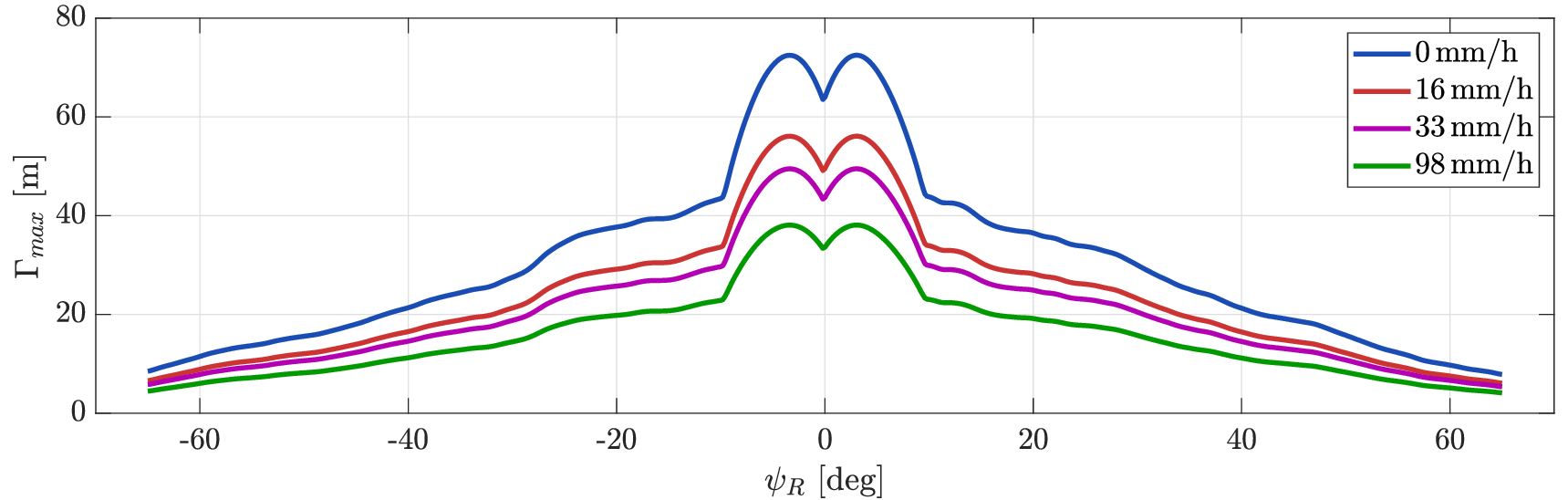}
  \captionsetup{font=footnotesize}
 	\caption{Radar detection range under different rain intensities and clear weather for a pedestrian target.}
 	\label{fig:thresholds_radar_rain_azimuth}
 \end{figure}
The simulated detection thresholds of the radar refer to $\psi_R = 0$° and are characterized by the sensor-specific and angle-dependent resolution for rain and fog in the respective Figures \ref{fig:thresholds_radar_rain_azimuth} and \ref{fig:thresholds_radar_fog_azimuth}. The obtained areas under the graphs of both plots represent the sensor's FOV. This angular constraint does not apply to the 360°-lidar used, so the related Figures \ref{fig:combined_lidar_radar_rain} and \ref{fig:combined_thresholds__radar_lidar_fog} apply for all $\psi_L \in [0,2\,\pi[$. Finally, Figures \ref{fig:sigma_radar} and \ref{fig:sigma_lidar} display the findings of our investigation into weather-related oscillations for each sensor across all measurement frames. Moreover, Figures \ref{fig:combined_lidar_radar_rain} and \ref{fig:combined_thresholds__radar_lidar_fog} illustrate that the WF prediction is more accurate than the baseline model in the tested scenarios.

\section{Discussion}\label{sec:discussion}
As outlined in Section \ref{sec:results_measurements}, measurement results (Figure \ref{fig:detection_points_radar} - \ref{fig:sigma_lidar}) confirm previous findings in the literature, indicating that the outcomes may not be caused by the individual environment and measurement conditions but represent the physical effects of rain and fog on object detection performance. While not being tested across the entire range, the WF showed a significantly more realistic behavior than the baseline model (see Figure \ref{fig:combined_lidar_radar_rain} and \ref{fig:combined_thresholds__radar_lidar_fog}). We believe this is sufficient evidence that the WF outperforms the baseline state-of-the-art model, even though the tuning effort is minimal. Further, given that the methodology can incorporate the physical properties of different target types, e.g., bicyclists and vehicles, and different relative angles, we conclude that such an extension would also be feasible. Similarly, the predictions of the WF could also apply to unseen scenarios. Such would indicate that the proposed method, i.e., the WF, allows for generalization to unseen scenarios due to the mixture of physical sensor representation and minimal empirical tuning.

\section{Conclusion and Future Work}\label{sec:conclusion}
This article examines the effects of fog and rain on pedestrian detection with lidar and radar sensors. Reproducible test data is scarce in the literature, and many state of the art models only represent the primary weather effects, i.e., siganl attenuation  (Section \ref{sec:introduction}).
Therefore, a model for the composition of signal attenuation due to fog and rain is given in Section \ref{sec:problem_formulation}. Further, empirical tuning coefficients are defined. For tuning such coefficients, comprehensive measurements are conducted to determine the overall effects of rain and fog on pedestrian detection with radar and lidar sensors (see Section \ref{sec:measurements}). On this basis, we propose the \textit{Weather Filter} (WF) in Section \ref{sec:weather_filter}, which combines a state-of-the-art baseline model with empirical data to constitute a model that includes both primary attenuation caused by aerosols and secondary weather effects, such as droplets on the sensor and changes in the reflectivity of the target. The results (Section \ref{sec:results}) show that the observations from the measurements agree with the literature related to weather degredation and that the WF predicts the effect of fog and rain on pedestrian detection with higher accuracy as a baseline state-of-the-art model. Also, it showed sufficient to tune the WF with a minimal number of measurements to approximate the behavior of radar and lidar under different rain and fog conditions. This leads to potential advantages in reducing the number of real test scenarios and, thus, the time and costs involved. To summarise, the WF model represents an important preliminary step towards enhancing the safety of automated vehicles in a diverse range of weather conditions by providing vital information about expected detection limits. This data enables the vehicle to adjust subsequent decision-making accordingly, thereby ensuring better safety. This work emphasizes the importance and needs for further research in this area. Future research will include other types of vulnerable road users and additional weather conditions, such as snow or the effect of water spray, along with a broader assessment of the WF's performance.

\section*{Acknowledgments}
This project has received funding from the European Union’s Horizon 2020 research and innovation program under Grant Agreement 861570, project SAFE-UP (proactive SAFEty systems and tools for a constantly UPgrading road environment).

\bibliography{bibliography.bib}
\bibliographystyle{IEEEtran}
\end{document}